\documentclass[pdflatex,sn-mathphys]{sn-jnl}

\jyear{2023}%

\theoremstyle{thmstyleone}%
\theoremstyle{thmstyletwo}%

\theoremstyle{thmstylethree}%

\begin{document}

\title[Article Title]{Deep Temporal Modelling of Clinical Depression through Social Media Text}


\author*[1]{\fnm{} \sur{Nawshad Farruque}}\email{nawshad@ualberta.ca}

\author[1]{\fnm{} \sur{Randy Goebel}}\email{rgoebel@ualberta.ca}

\author[2]{\fnm{} \sur{Sudhakar Sivapalan}}\email{sivapala@ualberta.ca}

\author[1]{\fnm{} \sur{Osmar R. Za\"{i}ane}}\email{zaiane@ualberta.ca}

\affil*[1]{\orgdiv{Department of Computing Science}, \orgname{Alberta Machine Intelligence Institute (AMII), Faculty of Science, University of Alberta}, \orgaddress{\city{Edmonton}, \postcode{T6G 2E8}, \state{AB}, \country{Canada}}}

\affil[2]{\orgdiv{Department of Psychiatry}, \orgname{Faculty of Medicine and Dentistry, University of Alberta}, \orgaddress{\city{Edmonton}, \postcode{T6G 2B7}, \state{AB}, \country{Canada}}}


\abstract{We describe the development of a model to detect user-level clinical depression based on a user's temporal social media posts. Our model uses a Depression Symptoms Detection (DSD) classifier, which is trained on the largest existing samples of clinician annotated tweets for clinical depression symptoms. We subsequently use our DSD model to extract clinically relevant features, e.g., depression scores and their consequent temporal patterns, as well as user posting activity patterns, e.g., quantifying their ``no activity'' or ``silence.'' Furthermore, to evaluate the efficacy of these extracted features, we create three kinds of datasets including a test dataset, from two existing well-known benchmark datasets for user-level depression detection. We then provide accuracy measures based on single features, baseline features and feature ablation tests, at several different levels of temporal granularity. The relevant data distributions and clinical depression detection related settings can be exploited to draw a complete picture of the impact of different features across our created datasets. Finally, we show that, in general, only semantic oriented representation models perform well. However, clinical features may enhance overall performance provided that the training and testing distribution is similar, and there is more data in a user's timeline. The consequence is that the predictive capability of depression scores increase significantly while used in a more sensitive clinical depression detection settings.}


\maketitle
\begin{abstract}


We describe the development of a model to detect user-level clinical depression based on a user's temporal social media posts. Our model uses a Depression Symptoms Detection (DSD) classifier, which is trained on the largest existing samples of clinician annotated tweets for clinical depression symptoms. We subsequently use our DSD model to extract clinically relevant features, e.g., depression scores and their consequent temporal patterns, as well as user posting activity patterns, e.g., quantifying their ``no activity'' or ``silence.'' Furthermore, to evaluate the efficacy of these extracted features, we create three kinds of datasets including a test dataset, from two existing well-known benchmark datasets for user-level depression detection. We then provide accuracy measures based on single features, baseline features and feature ablation tests, at several different levels of temporal granularity. The relevant data distributions and clinical depression detection related settings can be exploited to draw a complete picture of the impact of different features across our created datasets. Finally, we show that, in general, only semantic oriented representation models perform well. However, clinical features may enhance overall performance provided that the training and testing distribution is similar, and there is more data in a user's timeline. The consequence is that the predictive capability of depression scores increase significantly while used in a more sensitive clinical depression detection settings.

\end{abstract}

\section{Introduction}

Most of the earlier research in the area of user-level depression modelling through social media posts do not attempt to align with the clinical framework of depression detection. By clinical framework, we mean conforming to the definition of clinical depression as defined in DSM-5 \footnote{\url{https://www.psychiatry.org/psychiatrists/practice/dsm/feedback-and-questions/frequently-asked-questions}}, i.e. looking for signs of depression in at least a two-week episode of a user. Developing such a model is very challenging because it requires a Depression Symptoms Detection (DSD) model and a framework to calculate depression scores over the temporal episodes in a user's social media timeline. In this work, we mainly focus on using our learned DSD model and clinical insights for depression detection for extracting depression scores. We subsequently represent a user's timeline as a temporal series of depression scores then use that representation for our deep Temporal model of User-level clinical Depression (TUD). 

According to earlier research, social media posting activity patterns and language specific clues are very important for user-level depression modelling. Most current research has focused on these features in a non-temporal manner, i.e., on digests of tweets, meaning, taking all the tweets of a user's timeline and concatenate them together to represent that user, where temporal sequence of these tweets were not considered \cite{Choudhury2013Pred,coppersmith2015clpsych,Coppersmith2014,yazdavar2017semi,nguyen2022improving}. Very recently, a very closely related work was carried out by \cite{nguyen2022improving}, who inferred the presence of depression symptoms from each Reddit posts of a user. They extracted summaries of symptoms from arbitrary number of posts through different kernel size of a CNN classifier  and use those as non-temporal feature representations for user-level depression detection. Their depression presence calculation is based on looking for hand crafted text patterns of depression symptoms in a Reddit post. Relatively little research has considered temporal modelling. but showed a different focus instead of depression detection, e.g., finding correlation between depression score of the patients found through depression rating scales and their underlying mood patterns through social media text \cite{lushi2020examining} or tracking change of the same before the date of depression diagnosis \cite{reece2017forecasting}. Recently, \cite{zogan2022explainable} proposed a multi-modal depression detection from social media model which uses a hierarchical attention layer that leverages each tweets to learn word level and tweet level compositions. The main criticism of most of this research is about the value of extracted features: they fail to follow clinical depression modelling criteria and are primarily based on topical and lexical representations which are not as clinically useful as the clinical representations, i.e., depression scores. In addition temporal patterns analysis is missing.

Unlike earlier research, we extract depression scores for each of the two-week depressive episodes in a user's timeline and provide it to the temporal deep learning model, thus enabling the consideration of temporal modelling for user-level clinical depression. We also integrate user posting activity patterns through the proportion of the number of days they have posting activities out of all the days in an episode. This helps us distinguish between an episode without any signs of depression and the same period with no activity.  Earlier research was also not concerned about varying levels of granularity in a user's timeline. In our approach, we provide our model with a sliding a two-week time window of possible depression episode with various sliding lengths over a user's social media timeline, e.g., sliding length of 1, 7 and 14 days. In addition, and absent in the earlier research, we consider two different kinds of important depression modelling strategies: one follows strictly the clinical definition of depression, i.e., there must be social media posts that carry signs of either ``Anhedonia'' or ``Low mood'' in an episode to qualify it as an episode of depression; and the other does not. Since depression scoring depends on the thresholds used by the clinicians for determining whether a depression symptom is expressed either ``not at all,'' ``for several days,'' ``more than half of the days,'' or ``nearly everyday,''  we experiment with a more sensitive threshold, that qualifies an episode to be expressing depression even a user has exhibited a symptom in at least a day in that episode.  Therefore, the main motivation of this work comes from user level clinical depression modelling, which means, following clinical criteria of depression detection as laid out in DSM-5 and in clinical practice \footnote{\url{https://www.psychiatry.org/patients-families/depression/what-is-depression}}.

\section{Methodology}
\label{sec:tud-method}

We begin with an extensive analysis of our datasets. 
To do so, we first report distributions of different user specific statistics related to social media usage behavior, demography and linguistic components analysis based on a well-known psycholinguistic lexicon named, LIWC \cite{pennebaker2015development}. Next, we describe different clinical features based on depression scores and social media usage behavior of the users and how we extract them from our datasets. Later, we describe these feature distributions across our datasets. We then describe our deep learning model followed by the experimental setups, where we describe our sets of feature-ablated models and single-feature models compared to all-feature model and relevant baselines. 

We experiment with three types of depression episode analysis, starting from most granular to least granular. To do this, we slide a two-week temporal window in a user's time line from their earliest post in the history to the latest. We experiment with various slide length = (1, 7, 14). Slide length=1 provides us with the most granular temporal analysis to slide length=14 which is the least granular settings of the same. We keep the temporal window as two weeks to conform with the DSM-5 criteria of depression detection which defines depressive episodes to be of two weeks long. Moreover, it is found that temporal mood patterns are best captured through a two-week time window \cite{lushi2020examining}, and weekly windows are better than per-day analysis \cite{reece2017forecasting}. 

We experiment with two kinds of clinical depression detection settings; one strictly follows the clinical definition of depression and the other does not. We also experiment with two kinds of clinical analyses based on two different depression scoring strategies, one reflecting traditional clinical scoring approach, another reflecting more sensitive approach for depression detection. We create three main datasets for training purpose and separate a portion of each  for testing the performance of the model. We also create a separate test set from one of the datasets which is annotated for ongoing depressed users and then evaluate all the models in that set.


Finally, we provide detailed analysis on how different clinical features contribute to the user-level depression detection task in each of those datasets across various level of granularity and clinical settings.

\section{Datasets} 
\label{sec:tud-datasets}
We have created balanced data subsets from the CLPsych-2015 and IJCAI-2017 datasets  \cite{shen2017depression,coppersmith2015clpsych}. Both of these datasets are from Twitter users who self-disclosed their diagnosis of depression through a self-disclosing statement. In both of these datasets, depressed users are identified from Twitter users' self-disclosure and control users are the users who do not have such disclosures. We use balanced depressed and control subsets of users for our experiments, as it is found to be the most effective strategy to build robust user-level depression detection model by \cite{shen2017depression}, the curators of the largest benchmark dataset (IJCAI-2017 dataset) for the same task.

CLPsych-2015 users have markedly longer tweets history compared to IJCAI-2017 users. Moreover, IJCAI-2017 users have data preceding only one month of their self-disclosure. So analyzing IJCAI-2017 data in contrast to CLPsych-2015 provides a clear idea whether recency of self-disclosure has any effect on temporal user-level depression detection. In addition, our experiments are heavily based on social media posts from Twitter instead of Reddit or any other depression forums alike. The reason is that, we would like to use an unbiased representative of social media text, as opposed to using the datasets which have strong self-reporting bias such as depression forums. 

\subsection{Experiment Datasets Creation}
\label{subsec:exp-dataset-create}
We run experiments on three datasets. As described earlier, these datasets are extracted from two publicly available datasets: CLPsych-2015 and IJCAI-2017,
which are similar to most of the datasets previously reported: they use use public social media posts from self-disclosing users (i.e., Twitter users) for their depression condition. We describe the curation of these datasets as follows:

\subsubsection{CLPsych-2015-Users Dataset}
\label{subsubsec:clpsych-2015-users}

CLPsych-2015-Users dataset is a balanced subset of the CLPsych-2015 dataset. We ensure each user has minimum 50 posts and 30 days of Twitter history. This dataset does not include any self-disclosing statements. The original dataset from Twitter was created from users with the disclosure statement ``I was just diagnosed with depression''. Further, the original dataset creators employed human annotators to verify the authenticity of these self-disclosing statements for most of the users in that dataset. In addition, for a control population, random users were selected without such disclosing statements. The timeline for this dataset collection is in between the years 2008 and 2013.

\subsubsection{IJCAI-2017-Users Dataset}
\label{subsubsec:ijcai-2017-users}

We use a subset of the IJCAI-2017 dataset 
with users who have minimum 50 posts and 30 days of Twitter history. 
Note further, this is a multi-lingual dataset with users producing Tweets in different languages. To initially avoid the need for multi-lingual analysis, we discard user records who have more than 20\% non-English tweets. Even with this filter, we still find close to 1000 users. This dataset does include the self-disclosure statements from the users. For this dataset, the self-disclosure looks like the following text: ``I (am/was/have been) diagnosed with depression.'' Many of these disclosures also include the exact time of such diagnosis. Control users are identified based on the Twitter users who do not have any tweets with the character string ``depress.'' Because the Twitter API could return a huge number of tweets, the curators of this dataset restricted their collection of control tweets from the month of December, 2016. The timeline for collecting IJCAI-2017 depressed users is in between the years 2009 and 2016. Note further that this dataset contains the most recent one month of Tweets from the disclosure for depressed users; for control, it is just the recent one months of posts. 

Since for this dataset we have self-disclosure statement and the timeline of depression diagnosis, by analyzing each user's self-disclosure, we identify genuine users and two types of user datasets based on the recency of their diagnosis:

\begin{enumerate}
    \item \textbf{IJCAI-2017-Ongoing-Users}: these users declared that their depression diagnosis is recent. 
    \item \textbf{IJCAI-2017-Today-Users}: these users declared that they were diagnosed with depression exactly at the day of the disclosure.
\end{enumerate}

We identify genuine users based on the criteria that the user is talking about their own depression and those are not sarcasm, song lyrics or any other text that do not directly indicates a user's depression diagnosis. Whenever a user expressed any doubt about their depression diagnosis, we also consider them as not genuine. Details on the annotation task for finding out users with current/ongoing depression is provided in the work by \cite{macavaney2018rsdd}. We find only about 20\% of our IJCAI-2017 users as genuine ongoing depression candidate users. Moreover, we find only 9\% of those users who disclosed their exact date of depression diagnosis.

\subsubsection{Mixed-Users Dataset} 
\label{subsubsec:mixed-users}
We create a Mixed-Users dataset by combining both the above datasets to see whether combining both datasets would help in depression detection. We do not separately report the details of the combined dataset (e.g., feature and linguistic analysis) since it is the aggregate of our two main training datasets, i.e., CLPsych-2015-Users and IJCAI-2017-Ongoing-Users.

The choice of minimum number of posts, days and proportion of non-English tweets to curate for the above datasets is largely influenced by an earlier research which created one of the well-known benchmark datasets for user-level depression detection through Twitter timeline \cite{coppersmith2015clpsych}. Those authors selected users with a maximum 25\% non-English tweets and minimum 25 posts; we adopted a more restrictive strategy for non-English tweets proportion (i.e., 20\%) and minimum number of posts (i.e., 50) to facilitate more data per user. Our intuition is that these datasets will produce better models.

\subsection{Dataset Statistics}
\label{subsec:tud-dataset-stats}
Here we provide user-level social media behavior statistics including users' demographic profile.  We also provide linguistic component distribution analysis for the above mentioned datasets. For this linguistic analysis, we use a well-known psycholinguistic lexicon named, LIWC \cite{pennebaker2015development}, which we use to identify user-level mood fluctuation, emotion and sentiment analysis in temporal social media data. 
LIWC calculates the percentage of words in a text blurb arising from different psycholinguistic dimensions.

We do not have demographic information For the IJCAI-2017 dataset, nor do we have any information of the geographic location of the users. For the CLPsych-2015 dataset only, we have demographic information available. 

\subsubsection{User Specific Statistics:}
\label{subsubsec:tud-user-stats}
In our dataset statistics tables, we provide the following user specific statistics:

\begin{enumerate}
    \item \textbf{\#Users:} Total number of users.
    
    \item \textbf{Avg. Frequency. of Posting (AFP):} Time difference between two consecutive user activities; here activity means Tweet post by a user. AFP is the average of these differences. The lower the number, the more the activity or posting frequency of the user. 
    
    \item \textbf{Fluctuation of Posting Frequency (FPF):}  This is standard deviation of AFP, which approximates irregularity of a user's posting frequency.
    
    \item \textbf{\#Tweets:} Total number of tweets in a user's profile.
    
    \item \textbf{\#Proper-Tweets:} Total number of proper Tweets, i.e., Tweets after preprocessing all Tweets in  a user's timeline.
    
    \item \textbf{\#Days:} Total number of days a user has Twitter history.
    
    \item \textbf{Age:} Age of a user. Only available for CLPsych-2015 dataset, inferred by a third party machine learning model for detecting age \cite{coppersmith2015clpsych}.
    
    \item \textbf{Gender:} Gender of a user. Only available for CLPsych-2015 dataset, inferred by a third party machine learning model for detecting gender \cite{coppersmith2015clpsych}.
    
    \item \textbf{Avg. Tweets Length:} We report the average length of Tweets, i.e., average number of tokens in all Tweets in a user's timeline.
    
     \item \textbf{Avg. Sents :} We also report the average number of sentences in a Tweet. This is done by simply split a tweet based on period/question mark/exclamation.

\end{enumerate}

For all these statistics, we report the average and standard deviations across depressed and control population except \#Users (Tables \ref{tab:clpsych-2015-demographics} and \ref{tab:user-posting-stats}). We use Welch's two-tailed unpaired t-test to find statistical significance amongst the means of these features across depressed vs. control population (statistically significant means p-value $< 0.05$). Welch's unpaired t-test is a widely used method for comparing means between two populations \cite{gans1981use}. 

We observe that IJCAI-2017-Ongoing/Today-Users datasets are smaller than CLPsych-2015-Users: the average number of posts in CLPsych-2015-Users is higher in their timeline compared to IJCAI-2017-Ongoing/Today-Users.  However, average Tweet length and average number of sentences are the same across these datasets (Table \ref{tab:tud-dataset-stats}). 

In IJCAI-2017-Ongoing/Today-Users, the number of posts for the control population is higher than the depressed population, however, there is no such difference for the same in CLPsych-2015 users. \#Tweets and \#Proper-Tweets are significantly higher in the control dataset than the depressed Tweets in the IJCAI-2017-Ongoing/Today-Users. In CLPsych-2015 there is no such difference. In all three datasets, the average Tweet length is significantly higher in depressed population compared to the control population.

For both depressed and control CLPsych-2015-Users, we find there are more females than males and most of them are young adults (Table \ref{tab:clpsych-2015-demographics}), with the control population significantly older than the depressed population, by 4 years.

The Twitter timeline of CLPsych-2015-Users is significantly longer than IJCAI-2017-Ongoing/Test-Users. Moreover, in the CLPsych-2015-Users dataset, control users have significantly longer timelines than depressed users. For the IJCAI-2017-Ongoing/Test-Users, they are same, because, IJCAI-2017-Ongoing/Test Users datasets are collected for a window of 1 month only. 
In the CLPsych-2015-Users, depressed users post more frequently and show less fluctuation than control users; it's just the opposite for IJCAI-2017-Ongoing/Test Users. However, in both datasets, both control and depressed users are very active which is reflected through their AFP which is less than two days (Table \ref{tab:user-posting-stats}).


\begin{sidewaystable}
    \footnotesize
    \centering
    \begin{tabular}{|p{1cm}|p{1cm}|p{1cm}|p{1cm}|p{1cm}|p{1.2cm}|p{1.2cm}|p{1.2cm}|p{1.2cm}|p{1.2cm}|p{1.2cm}|}
    \hline
        \textbf{Datasets} & \textbf{Sample-Size (D)} & \textbf{Sample-Size (C)} & \textbf{\#Tweets (D)} & \textbf{\#Tweets (C)} & \textbf{Avg. Tweets Length(D)} & \textbf{Avg. Tweets Length(C)} & \textbf{Avg. Sents(D)} & \textbf{Avg. Sents(C)} & \textbf{\#Proper-Tweets (D)} & \textbf{\#Proper-Tweets (C)} \\ \hline
        CLPsych-2015-Users & 273 & 264 & $1067$ & $1044$ & $13.2^*$ & $12.4$ & $1.5$  & $1.6^*$  &  $1020$ & $999$ \\ \hline
        IJCAI-2017-Ongoing-Users & 196 & 196 & $187$ & $425^*$ & $13.8^*$ & $12.5$ & $1.5^*$  & $1.4$  & $167$ & $372^*$ \\ \hline
        IJCAI-2017-Today-Users & 18 & 18 & $155$ & $362^*$ & $14.6^*$ & $12.1$ & $1.6$  & $1.5$  &  $142$ & $292^*$ \\ \hline
    \end{tabular}
    \caption{Dataset statistics for all datasets (* indicates significantly higher with p-value $< 0.05$ in Welch's two-tailed unpaired t-test). \label{tab:tud-dataset-stats}}
\end{sidewaystable}

\begin{table}
    \centering
    \begin{tabular}{|p{2cm}|p{2cm}|p{2cm}|p{2.5cm}|}
    \hline
        \textbf{Class} & \textbf{\#Male} & \textbf{\#Female} & \textbf{Age (Mean)} \\ \hline
        Control & 74 & 190 & $25.2^*$ \\ \hline
        Depression & 54 & 219 & $21.6$ \\ \hline
    \end{tabular}
    \caption{CLPsych-2015 demographic statistics (* indicates significantly higher with p-value $< 0.05$ in Welch's two-tailed unpaired t-test). \label{tab:clpsych-2015-demographics}}
\end{table}

\begin{table}
    \footnotesize
    \centering
    \begin{tabular}{|p{2cm}|p{2cm}|p{2cm}|p{0.6cm}|p{0.6cm}|p{0.6cm}|p{0.6cm}|}
    \hline
        \textbf{Datasets} & \textbf{\#HistoryDays (D)} & \textbf{\#HistoryDays (C)} & \textbf{AFP (D)} & \textbf{AFP (C)} & \textbf{FPF (D)} & \textbf{FPF (C)} \\ \hline
        CLPsych-2015-Users & $366$ & $495^*$ & $0.62$ & $1.139*$ & $3.18$ & $5.04^*$ \\ \hline
        IJCAI-2017-Ongoing-Users & $30$ & $30$ & $0.09^*$ & $0.04$ & $0.35^*$ & $0.17$ \\ \hline
        IJCAI-2017-Today-Users & $30$ & $30$ & $0.13^*$ & $0.05$ & $0.49$ & $0.23$ \\ \hline
    \end{tabular}
    \caption{User posting related statistics for all datasets (* indicates significantly higher with p-value $< 0.05$ in Welch's two-tailed unpaired t-test). \label{tab:user-posting-stats}}
\end{table}

\subsubsection{Linguistic Components Distribution}
\label{subsubsec:tud-ling-comp}

Here we provide the linguistic component analysis with the help of LIWC. We first create a digest of all Twitter posts from both depressed and control users' Twitter timelines. We then apply LIWC on this digest. LIWC finds the proportion percentage of lexicon items under each lexicon components, which we call lexicon component intensity (LCI). We follow the steps provided below to perform our linguistic component distribution analysis:

\begin{enumerate}
    \item We find the deviations between the LCIs (we call $LCI_{dev}$) for depression ($LCI_{d}$) and control ($LCI_{c}$) population for each dataset. All the positive values (or deviations) mean those components have high $LCI$ in depressed population compared to the control population; negative means vice-versa, and zero means equal (Equation \ref{eq:lci-dev}).
    
    \item Finally, we report $LCI_{dev}$ for all the common LIWC components where $LCI_{dev} > 0$ for the depressed population and the control population for all three datasets. For the control population, we make negative deviation positive. We then report the average and standard deviation of those in the Tables \ref{tab:depression-deviations} and \ref{tab:non-depression-deviations} in descending order of the average  $LCI_{dev}$ across all datasets (Equations \ref{eq:lci-dev}, \ref{eq:lci-dev-avg} and \ref{eq:lci-dev-std.dev.}) for the calculation of these measures. 
    
    This analysis provides us with the LIWC components that are most clearly expressed in depressed population compared with the control population and vice versa.
    
    \begin{equation}
    \label{eq:lci-dev}
        LCI_{dev} =
            LCI_{d} - LCI_{c} 
    \end{equation}
    
    \begin{equation}
    \label{eq:lci-dev-avg}
       LCI_{dev-avg} = \mu(LCI_{dev})
    \end{equation}
    
    \begin{equation}
    \label{eq:lci-dev-std.dev.}
       LCI_{dev-std} = \sigma(LCI_{dev})
    \end{equation}
    
\end{enumerate}


\begin{table}
    \centering
    \begin{tabular}{|p{2cm}|p{1.5cm}|p{1.5cm}|p{1.5cm}|p{1.5cm}|p{2cm}|}
    \hline
        \textbf{LIWC Components} & \textbf{CLPsych-2015} & \textbf{IJCAI-2017-Ongoing-Users} & \textbf{IJCAI-2017-Today-Users} & $\mathbf{LCI_{dev-avg}}$ & $\mathbf{LCI_{dev-std.dev.}}$ \\ \hline
        Authentic & 13.74 & 7.40 & 7.61 & 9.58 & 3.60 \\ \hline
        Linguistic & 8.1 & 3.50 & 0.29 & 3.96 & 3.93 \\ \hline
        function & 7.61 & 2.97 & 0.97 & 3.85 & 3.41 \\ \hline
        Dic & 6.17 & 3.74 & 0.97 & 3.63 & 2.60 \\ \hline
        \textbf{i} & 2.92 & 1.16 & 1.65 & 1.91 & 0.91 \\ \hline
        \textbf{pronoun} & 3.93 & 1.46 & 0.23 & 1.87 & 1.88 \\ \hline
        \textbf{ppron} & 3.27 & 1.24 & 0.39 & 1.63 & 1.48 \\ \hline
        Cognition & 1.9 & 1.55 & 0.30 & 1.25 & 0.84 \\ \hline
        cogproc & 1.68 & 1.45 & 0.54 & 1.22 & 0.60 \\ \hline
        auxverb & 2.08 & 0.50 & 0.24 & 0.94 & 1.00 \\ \hline
        conj & 1.37 & 0.81 & 0.46 & 0.88 & 0.46 \\ \hline
        Period & 0.07 & 1.36 & 1.14 & 0.86 & 0.69 \\ \hline
        adverb & 1.37 & 0.43 & 0.51 & 0.77 & 0.52 \\ \hline
        \textbf{emotion} & 0.57 & 0.51 & 0.60 & 0.56 & 0.05 \\ \hline
        focuspresent & 1.14 & 0.12 & 0.20 & 0.49 & 0.57 \\ \hline
        \textbf{tone\_neg} & 0.49 & 0.44 & 0.33 & 0.42 & 0.08 \\ \hline
        \textbf{emo\_neg} & 0.39 & 0.37 & 0.32 & 0.36 & 0.04 \\ \hline
        \textbf{health} & 0.34 & 0.65 & 0.05 & 0.35 & 0.30 \\ \hline
        insight & 0.16 & 0.15 & 0.48 & 0.26 & 0.19 \\ \hline
        cause & 0.19 & 0.27 & 0.23 & 0.23 & 0.04 \\ \hline
        \textbf{emo\_pos} & 0.18 & 0.13 & 0.24 & 0.18 & 0.06 \\ \hline
        \textbf{emo\_sad} & 0.11 & 0.07 & 0.20 & 0.13 & 0.07 \\ \hline
        certitude & 0.16 & 0.05 & 0.17 & 0.13 & 0.07 \\ \hline
        \textbf{illness} & 0.07 & 0.28 & 0.02 & 0.12 & 0.14 \\ \hline
        \textbf{mental} & 0.16 & 0.14 & 0.03 & 0.11 & 0.07 \\ \hline
        family & 0.06 & 0.16 & 0.10 & 0.11 & 0.05 \\ \hline
        want & 0.13 & 0.14 & 0.04 & 0.10 & 0.06 \\ \hline
        \textbf{feeling} & 0.12 & 0.08 & 0.11 & 0.10 & 0.02 \\ \hline
        assent & 0.06 & 0.08 & 0.03 & 0.06 & 0.03 \\ \hline
        friend & 0.05 & 0.04 & 0.08 & 0.06 & 0.02 \\ \hline
        sexual & 0.06 & 0.07 & 0.01 & 0.05 & 0.03 \\ \hline
        \textbf{emo\_anx} & 0.08 & 0.04 & 0.01 & 0.04 & 0.04 \\ \hline
        lack & 0.02 & 0.02 & 0.01 & 0.02 & 0.01 \\ \hline
    \end{tabular}
    \caption{Depression deviations for all three datasets\label{tab:depression-deviations}}
\end{table}

\begin{table}
    \centering
    \begin{tabular}{|p{2cm}|p{1.5cm}|p{1.5cm}|p{1.5cm}|p{1.5cm}|p{2cm}|}
    \hline
        \textbf{LIWC Components} & \textbf{CLPsych-2015-Users} & \textbf{IJCAI-2017-Ongoing-Users} & \textbf{IJCAI-2017-Today-Users} & $\mathbf{LCI_{dev-avg}}$ & $\mathbf{LCI_{dev-std.dev.}}$ \\ \hline
        Analytic & 27.89 & 8.93 & 2.14 & 12.99 & 13.35 \\ \hline
        Clout & 17.88 & 5.89 & 9.66 & 11.14 & 6.13 \\ \hline
        Tone & 12.63 & 12.41 & 7.35 & 10.80 & 2.99 \\ \hline
        Lifestyle & 1.22 & 0.34 & 0.14 & 0.57 & 0.57 \\ \hline
        Perception & 0.71 & 0.46 & 0.50 & 0.56 & 0.13 \\ \hline
        netspeak & 0.13 & 0.31 & 0.91 & 0.45 & 0.41 \\ \hline
        Drives & 0.81 & 0.03 & 0.42 & 0.42 & 0.39 \\ \hline
        space & 0.53 & 0.24 & 0.38 & 0.38 & 0.15 \\ \hline
        Conversation & 0.04 & 0.24 & 0.79 & 0.36 & 0.39 \\ \hline
        food & 0.29 & 0.07 & 0.45 & 0.27 & 0.19 \\ \hline
        leisure & 0.35 & 0.24 & 0.07 & 0.22 & 0.14 \\ \hline
        tone\_pos & 0.26 & 0.29 & 0.11 & 0.22 & 0.10 \\ \hline
        Culture & 0.31 & 0.26 & 0.02 & 0.20 & 0.16 \\ \hline
        power & 0.26 & 0.04 & 0.28 & 0.19 & 0.13 \\ \hline
        motion & 0.17 & 0.06 & 0.31 & 0.18 & 0.13 \\ \hline
        we & 0.24 & 0.26 & 0.01 & 0.17 & 0.14 \\ \hline
        affiliation & 0.31 & 0.04 & 0.04 & 0.13 & 0.16 \\ \hline
        reward & 0.11 & 0.24 & 0.01 & 0.12 & 0.12 \\ \hline
        relig & 0.00 & 0.10 & 0.14 & 0.08 & 0.07 \\ \hline
        politic & 0.15 & 0.04 & 0.03 & 0.07 & 0.07 \\ \hline
        ethnicity & 0.03 & 0.01 & 0.03 & 0.02 & 0.01 \\ \hline
    \end{tabular}
    \caption{Control deviations for all three datasets\label{tab:non-depression-deviations}.}
\end{table}

These tables show that the language used by the depressed population has more use of personal pronouns, negative emotion and anxiety related words compared to the control population (bold items in Table \ref{tab:depression-deviations}). This observation aligns with earlier research, such as \cite{Choudhury2013Pred}.

\section{Data Preprocessing}
\label{sec:tud-data-preproc}
\subsection{Tweets Preprocessing}
We use the following data preprocessing for the Tweets.



\begin{enumerate}
    \item Lowercase each words.
    \item Remove re-tweets and replies.
    \item Remove one character words (except ``a'', ``i'' and ``u'') and digits.
    \item Remove tweets which are less than three words long.
    \item Re-contract contracted words in a tweet. For example, ``I've'' is made ``I have''.
    \item Elongated words are converted to their original form. For example, ``Looong'' is turned to ``Long''.
    \item Remove tweets with self-disclosure, i.e. any tweet containing the word ``diagnosed'' or ``diagnosis'' is removed.
    \item Remove all punctuation except period, comma, question mark and exclamation. Punctuation have been found useful to represent a text based on sentence embedding.
    \item Remove URLs.
    \item Remove non-ASCII characters from words.
    \item Remove hashtags
    \item Remove emojis.
\end{enumerate} 

All the Tweets which are excluded after preprocessing are counted towards user posting activity but they don't carry signs of depression. ``No posting activity'' or absence is represented differently than absence of depression, so that our modelling can distinguish between these two.

\subsection{User Level Filtering}
Our datasets are derived from two widely used benchmark datasets used by numerous established research \cite{Coppersmith2014,jamil2017monitoring,orabi2018deep,shen2017depression,yadav2020identifying} without any user filtering. One reason is because the original data curators already verified the users through human annotation and analysing the genuineness of their disclosure \cite{coppersmith2015clpsych,shen2017depression}. In addition, our own Tweets preprocessing and minimum 50 number of posts constraint also removes users with excessive gibberish and irregular users. Finally, we also manually reviewed each user's timeline to verify the quality of the users based on the content of their posts, i.e., whether they have at least a few posts regarding their struggles related to depression.

\section{Clinically Relevant Features Extraction}\label{sec:tud-clinical-feats-extract}
Here we describe how we calculate several clinically relevant features for an episode (i.e., for a two-week time window). We later use these features to learn temporal patterns using our Deep Temporal model of User-level clinical Depression (TUD).

\subsection{Depression Score (DS)} 
\label{sec:tud-dep-score}

One of the major contributions of our research is to employ the DSD model 
to guide extraction of depression scores for an episode. We extract such depression scores over all such episodes in a user's Twitter timeline and then use TUD to learn useful temporal patterns of depression. To develop a robust DSD model, we created a framework, where, we used DSD dataset curated by \cite{yadav2020identifying}. At the heart of our DSD model is a Mental-Health pre-trained BERT, which was further fine-tuned with the help of clinician annotated depression symptoms tweets. Later, we improved the model's accuracy through a semi-supervised learning framework, where, we harvested candidate tweets for clinical depression symptoms from depressed Twitter users' timelines and DSD dataset curated by Yadav et al. We then re-trained our model with this harvested dataset. We found a final DSD model which achieved significantly more accuracy than its initial version. Also, at the same time, we harvested more clinical symptoms annotated tweets through this process than the amount we started with. Both our clinician annotated depression symptoms annotated dataset and the harvested dataset are the largest of their kind compared to early work. More details of the DSD model creation is provided in the research work by \cite{farruque2022depression}.

To enable our feature extraction process we take the following steps:

\begin{enumerate}
    \item We first sort the posts of a user based on their Twitter post timestamp information, in an ascending order of recency.
    \item We then create day-wise chunks of Tweets.
    \item For each day of Tweet chunks we calculate a Depression Symptoms Expression Vector (DSEV), where $DSEV \in \{0,1\}^d$, and $d$ = \#depression-symptoms. Each index of this vector corresponds to each of the 10 depression symptoms we are interested in. DSEV is initialized as all 0s at the beginning, indicating no symptoms is expressed; then if any of the Tweets in the chunk has expressed symptoms,  a particular index of DSEV vector is assigned value 1, which signifies a corresponding depression symptom is expressed for that day (Details provided in Algorithm \ref{algo:dsev-vector}).
    
    \begin{algorithm}[H]
        \small
        \SetAlgoLined
        \KwIn{A day chunk of tweets, $D$, Depression symptoms, $S$}
        \KwOut{$DSEV$}
            $DSEV \leftarrow \{0\}^{\abs{S}}$ \;
            \ForEach{$tweet \in day$}{
                $depSymptsIDs \leftarrow DSD(tweet)$ \;
                \ForEach{$symptIndex \in range(\abs{S})$}{
                    \If{$symptIndex \in depSymptsIDs$}{
                        $DSEV[symptIndex] \leftarrow 1$
                    }
                    
                }
            }
        return $DSEV$ \;
        \caption{\label{algo:dsev-vector}Depression-Symptoms-Expression-Vector Algorithm(DSEVA)}
    \end{algorithm}

    \item Later, in the first layer of TUD, we extract all the DSEVs in an episode, aggregate them and calculate the percentage of days each depression symptoms is expressed (Lines (5-12) in Algorithm \ref{algo:dsa}). We calculate this percentage on the number of days the user has activity, i.e., Twitter posts.
    
    
     \begin{algorithm}[H]
        \small
        \SetAlgoLined
        \KwIn{An episode, $E$, depression symptoms, $S$, Mode, $M$}
        \KwOut{Depression Score, $depScore$}
         $depScore \leftarrow 0$ \;
         $symptomScore \leftarrow 0$ \;
         $DSEV_{Sum} \leftarrow 0$ \;
         $symptomScoreSum \leftarrow 0$ \;
         
         \ForEach{$day \in E$}{ 
            $DSEV_{sum} \leftarrow DSEV_{sum} + DSEVA(day, S)$
         }
        
        \ForEach{$symptIndex \in range(\abs{S})$}{
            $symptomScoreSum \leftarrow DSEV_{sum}[symptIndex]$
            $percentOfDays \leftarrow (symptomScoreSum/{\abs{E}}) \times 100$
            
            \If {$(percentOfDays \geq 50)$ \textbf{and} ($S$ \textbf{is} (``Anhedonia'' \textbf{or} ``Low Mood''))}{
                $isClinicallyDepressed \leftarrow True$
            }
            
            \If {$(percentOfDays \geq 0)$ \textbf{and} $(percentOfDays < 20)$}{
                $symptomScore \leftarrow 0$
            }
            \ElseIf {$(percentOfDays \geq 20)$ \textbf{and} $(percentOfDays < 50)$}{
                $symptomScore \leftarrow 1$
            }
            \ElseIf {$(percentOfDays \geq 50$) \textbf{and} $percentOfDays < 85 $}{
                $symptomScore \leftarrow 2$
            }
            \ElseIf {$(percentOfDays \geq 85) $}{
                $symptomScore \leftarrow 3$
            }
            $depScore \leftarrow  depScore + symptomScore$ \;
        }
        \If {$M$ \textbf{is} ``Clinical''}{
            \If {$isClinicallyDepressed$}{
                return $depScore$ \;
            }
            \Else{
                return 0 \;
            }
        }
        \ElseIf{$M$ \textbf{is} ``Non-Clinical''}{
            return $depScore$ \;
        }
        \caption{\label{algo:dsa}Depression-Score Algorithm (DSA).}
    \end{algorithm}
    
    \item A user may not have tweets in each episode for all of its days. 
    So we also keep track of the days for which a user has no activity (i.e. no Tweets), which we use to calculate an Absence-Ratio (AR). This is further discussed in Section \ref{subsec:tud-ap}.

    \item Finally, Depression Score (DS) is calculated based on the percentage of days for appearance of each depression symptom. Here we consider ``Agitation'' and ``Retardation'' as one symptom, instead of two separate ones to conform with PHQ-9 (for the sake of brevity this is not included in the algorithm). If this is within a predefined range of thresholds \footnote{It is to be noted that the predefined thresholds in PHQ-9 are not concrete; they are roughly described, so we use a clinician's advice to ground those descriptions to numerical values (Appendix \ref{appen:depression-level-mapping})}, as defined in PHQ-9, we assign a corresponding score (or symptomScore) for that symptom in an episode. Aggregating all these scores for all the symptoms provide us with the final depression score (Lines (5-26) in Algorithm \ref{algo:dsa}). 
    \item In order to identify clinical depression, a user must have either ``Low Mood'' or ``Anhedonia,'' so we adjust our scoring Algorithm \ref{algo:dsa} (Lines 27-37), so that we can calculate depression scores by fulfilling the clinical criteria or relaxing it. The former option is called \textbf{Clinical Scoring (CS)}, the latter is called \textbf{Non-clinical scoring (NCS)}. In CS criteria, a depression score of 0 is assigned for an episode, if none of the above depression symptoms are expressed in that episode; otherwise, we move on with the depression score calculation as described earlier. We report our TUD performance for both options. 
    
    \item We also consider a much more sensitive version of depression scoring. So instead of considering all these thresholds stated in lines (13-26) in Algorithm \ref{algo:dsa}, we consider only one threshold, i.e., whenever there is a Tweet carrying signs of depression, we consider a symptomScore of 1, and an episode will be considered as a mild depression episode whenever, it has $depScore > 0$, otherwise the episode will not be considered as a depression episode, we call this \textbf{Minimal Depression Expression (MDE)} based Temporal Modelling.
\end{enumerate}

\subsection{Semantic Information} \label{subsec:tud-sem-info}
To create a representation that captures semantic information corresponding to a depression episode, we first take the average of the sentence embeddings for all the Tweets in a day to represent that day. We call this \textbf{Day-Level-Sentence-Embedding-Average (DLSEA)}. Subsequently, based on this day level semantic representation, we calculate the episode level semantic representation by again calculating the average embedding for all the DLSEAs in an episode. We call this \textbf{Episode-Level-Sentence-Embedding-Average (ELSEA)}. We also take all the Tweets and the average of their sentence embeddings, which we call \textbf{All-Tweets-Embedding-Average(ATEA)}. We use a Universal Sentence Encoder (USE) based sentence embedding for all these representations, as USE has been found out to be an effective and compact representation for many NLP tasks \cite{cer2018universal}. 


\subsection{User Posting Activity Pattern} 
\label{subsec:tud-ap}
As mentioned earlier, we determine posting activity patterns for each episodes. To do so, we calculate the number of days a user has no activity (or no social media posts in a day) out of all days in an episode, which we call the \textbf{Absence-Ratio (AR)}. 


\subsection{Temporal Depression Patterns} \label{subsec:tud-tp}
We extract two kinds of temporal depression patterns among all the episodes with user activity. 
These are (1) Depression Recurrence Frequency and (2) Inertia. Depression Recurrence Frequency has been found to be an important predictor of clinical depression as it is usually highly recurrent in nature \cite{burcusa2007risk}, and Inertia has been found by early research as an important trait for depressed social media users \cite{kuppens2012emotional}.

To calculate those, we first binarize the temporal series of episodic depression scores. We call this series \textbf{Binarized Temporal Episodes (BTE)}. Through binarization we convert the depression scores to 1 if those correspond to minimal or higher level of depression, otherwise we convert it to 0 (Algorithm \ref{algo:btea}). We later take the following steps to calculate \textbf{Depression Recurrence Frequency} and \textbf{Inertia} scores.

    \begin{algorithm}[!h]
        \small
        \SetAlgoLined
        \KwIn{Temporal Episodes, $TE$}
        \KwOut{Binarized Temporal Episodes, $BTE$}
            $BTE \leftarrow \{0\}^{\abs{TE}}$ \;
            \ForEach{$E \in TE$}{
                \If{$depLevel(E) \geq ``MINIMAL''$}{
                    $BTE \leftarrow 1$
                }
            }
        return $BTE$ \;
        \caption{\label{algo:btea}Binarized-Temporal-Episodes Algorithm (BTEA).}
    \end{algorithm}

\subsubsection{Depression Recurrence Frequency Score (DRFS)} 
\label{subsubsec:tud-drfs}
Depression recurrence represents the repetition of depressive mood. Here, we track whether a user's depression shows up in a cyclic manner. To calculate this, we first compress BTE, or remove consecutive repetitive binary scores; we call this series \textbf{Compressed Binarized Temporal Episodes (CBTE)}. 
Later, we find the cycles of the pattern ``1-0-1'', which means, a user starts with depression, gets better but again falls into depression. We calculate all such cycle in CBTE and normalize that with the number of items or binary scores in CBTE. We call this score \textbf{DRFS} (Algorithms \ref{algo:cbtea}, \ref{algo:cca} and \ref{algo:drfsa}).

    \begin{algorithm}[!h]
        \small
        \SetAlgoLined
        \KwIn{Binarized-Temporal-Episodes, $BTE$}
        \KwOut{Compressed-Binarized-Temporal-Episodes, $CBTE$}
            $CBTE \leftarrow \emptyset$ \;
            $CBTE.insert(BTE[0])$ \;
            \ForEach{$TE \in BTE$}{
                \If{$TE \neq CBTE[\abs{CBTE}- 1]$}{
                    $CBTE.insert(TE)$
                }
            }
        return $CBTE$ \;
        \caption{\label{algo:cbtea}Compressed-Binarized-Temporal-Episodes Algorithm (CBTEA).}
    \end{algorithm}
    
    \begin{algorithm}[!h]
        \small
        \SetAlgoLined
        \KwIn{Compressed-Binarized-Temporal-Episodes, $CBTE$}
        \KwOut{Cycle-Count, $CC$}
        $CC \leftarrow 0$ \;
        
        \If{$\abs{CBTE} > 2$}{
            \ForEach{$index \in range(\abs{CBTE}$}{
                \If{$0 < index < (\abs{CBTE} - 1)$}{
                    \If{$CBTE[index] = 0$}{
                        \If{($CBTE[index-1] = 1)$ \textbf{and} $(CBTE[index+1] = 1)$}{
                            $CC \leftarrow CC + 1$
                        }
                    }
                }
            }
        }
        return $CC$ \;
        \caption{\label{algo:cca} Cycle-Count Algorithm(CCA).}
    \end{algorithm}

    \begin{algorithm}[!h]
        \small
        \SetAlgoLined
        \KwIn{Compressed-Binarized-Temporal-Episodes, $CBTE$}
        \KwOut{Depression-Recurrence-Frequency-Score, $DRFS$}
            $cycles \leftarrow CCA(CBTE)$ \;
            $DRFS \leftarrow (\abs{cycles} / \abs{CBTE})$ \;
        return $DRFS$ \;
        \caption{\label{algo:drfsa}Depression-Recurrence-Frequency-Score Algorithm (DRFSA).}
    \end{algorithm}

\subsubsection{Inertia Score (IS)}
\label{subsubsec:tud-is}
Inertia means the tendency of staying in depressive mood for some extended period of time (e.g., multiple two week periods). To calculate this, we take BTE and find how many consecutive episodes have values 1, which means how many consecutive depressive episodes are there in a user's timeline. We then normalize this count with the total episode counts of that user. We call this score \textbf{Inertia Score (IS)} (Algorithm \ref{algo:isa}).

    \begin{algorithm}[!h]
        \small
        \SetAlgoLined
        \KwIn{Binarized-Temporal-Episodes, $BTE$}
        \KwOut{Inertia-Score, $IS$}
            $consecutivenessCount \leftarrow 0$ \;
            \ForEach{$index \in range(\abs{BTE} - 1)$}{
                \If{$BTE[index] - BTE[index + 1] = 0$}{
                    \If{$BTE[index] \times BTE[index + 1] = 1$}{
                        $consecutivenessCount\leftarrow consecutivenessCount + 1$
                    }
                }
            }
        $IS \leftarrow (consecutivenessCount / \abs{BTE})$ \;
        return $IS$ \;
        \caption{\label{algo:isa}Inertia-Score Algorithm (ISA).}
    \end{algorithm}

\section{Clinically Relevant Feature Distribution in the Datasets}
\label{sec:tud-clinical-feats-distr}
Here we report the extracted feature distributions, such as depression levels, depression score related temporal 
patterns (i.e., DRFS and IS) and user-activity patterns for our three datasets (i.e., CLPsych-2015-Users, IJCAI-2017-Ongoing-Users and IJCAI-2017-Today-Users). To calculate this distribution, we first determine the proportion of episodes out of all the episodes in a user's Twitter timeline. We then find out the average and standard deviation of these measures for all the users in depressed and control population. These numbers are reported in Tables  \ref{tab:clpsych-2015-ongoing-feats-distr}, \ref{tab:ijcai-2017-ongoing-feats-distr} and \ref{tab:ijcai-2017-today-feats-distr}. We report differences among these features across depression versus control populations, based on Welch's two-tailed unpaired t-test (statistically significant means p-value $< 0.05$).


\begin{table}[!h]
    \centering
    \begin{tabular}{|l|l|l|}
    \hline
        \textbf{Depression-Level} & \textbf{Control} & \textbf{Depression} \\ \hline
        None                &$0.9763(\pm{0.1167})^*$  &$0.9258(\pm{0.1885})$ \\ \hline
        None(MDE)           &$0.6701(\pm{0.3405})^*$  &$0.4808(\pm{0.3281})$ \\ \hline
        Minimal             &$0.0125(\pm{0.0577})$  &$0.03649(\pm{0.0949})^*$ \\ \hline
        Minimal(MDE)        &$0.3299(\pm{0.3405})$  &$0.5192(\pm{0.3281})^*$ \\ \hline
        Mild                &$0.0111(\pm{0.0717})$      &$0.0375(\pm{0.1519})^*$ \\ \hline
        Moderate            &$0(\pm{0})$            &$0.0002(±0.0026)$ \\ \hline
        Moderately-Severe   &$0(\pm{0}$ &$0\pm{0})$ \\ \hline
        Severe              &$0(\pm{0})$ &$0(\pm{0})$ \\ \hline
        AR                  &$0.3616(\pm{0.2771})$ &$0.3600(\pm{0.2660})$ \\ \hline
        IS                  &$0.0211(\pm{0.1099})$ &$0.0664(\pm{0.1750})^*$ \\ \hline
        IS(MDE)             &$0.3160(\pm{0.3310})$ &$0.4996(\pm{0.3216})^*$ \\ \hline
        DRFS                &$0.0013(\pm{0.0068})$ &$0.0039(\pm{0.0105})^*$ \\ \hline
        DRFS(MDE)           &$0.0083(\pm{0.0105})$ &$0.0104(\pm{0.0098})^*$ \\ \hline
    \end{tabular}
    \caption{\label{tab:clpsych-2015-ongoing-feats-distr} CLPsych-2015-Users features distribution (* indicates significantly higher with p-value $< 0.05$ in Welch's two-tailed unpaired t-test).}
\end{table}

\begin{table}[!h]
    \centering
    \begin{tabular}{|l|l|l|}
    \hline
        \textbf{Depression-Level}   &\textbf{Control} &\textbf{Depression} \\ \hline
        None                        &$0.9271(\pm{0.2258})$  &$0.9232(\pm{0.2114})$ \\ \hline
        None(MDE)                   &$0.3328(\pm{0.4117})^*$  &$0.2290(\pm{0.3433})$ \\ \hline
        Minimal                     &$0.0396(\pm{0.1484})$  &$0.0528(\pm{0.1553})$ \\ \hline
        Minimal(MDE)                &$0.6671(\pm{0.4116})$  &$0.7710(\pm{0.3433})^*$ \\ \hline
        Mild                        &$0.0327(\pm{0.1624})$   &$0.024(\pm{0.1279})$ \\ \hline
        Moderate                    &$0.0006(\pm{0.0084})$    &$0(\pm{0})$ \\ \hline
        Moderately-Severe           &$0(\pm{0})$              &$0(\pm{0})$           \\ \hline
        Severe                      &$0(\pm{0})$              &$0(\pm{0})$                   \\ \hline
        AR                          &$0.0827(\pm{0.1177})$    &$0.1691(\pm{0.1585})^*$ \\ \hline
        IS                          &$0.0621(\pm{0.2063})$    &$0.0615(\pm{0.1878})$ \\ \hline
        IS(MDE)                     &$0.6164(\pm{0.3903})$    &$0.7113(\pm{0.3318})^*$ \\ \hline
        DRFS                        &$0.0021(\pm{0.015})$     &$0.0045(\pm{0.020})$ \\ \hline
        DRFS(MDE)                   &$0.0048(\pm{0.0161})$    &$0.0066(\pm{0.0186})$ \\ \hline
    \end{tabular}
    \caption{\label{tab:ijcai-2017-ongoing-feats-distr}IJCAI-2017-Ongoing-Users features distribution (* indicates significantly higher with p-value $< 0.05$ in Welch's two-tailed unpaired t-test).}
\end{table}

\begin{table}[!h]
    \centering
    \begin{tabular}{|l|l|l|}
    \hline
        \textbf{Depression-Level}   &\textbf{Control} &\textbf{Depression} \\ \hline
        None                        &$0.9935(\pm{0.0278})$ &$1(\pm{0})$ \\ \hline
        None(MDE)                   &$0.2549(\pm{0.3406})$  &$0.2974(\pm{0.3909})$ \\ \hline
        Minimal                     &$0.0065(\pm{0.0278})$ &$0(\pm{0})$ \\ \hline
        Minimal(MDE)                &$0.7451(\pm{0.3406})$  &$0.7026(\pm{0.3909})$ \\ \hline
        Mild                        &$0(\pm{0})$ &$0(\pm{0})$ \\ \hline
        Moderate                    &$0(\pm{0})$ &$0(\pm{0})$ \\ \hline
        Moderately-Severe           &$0(\pm{0})$ &$0(\pm{0})$ \\ \hline
        Severe                      &$0(\pm{0})$ &$0(\pm{0})$ \\ \hline
        AR                          &$0.1127(\pm{0.1902})$ &$0.2374(\pm{0.2293})$ \\ \hline
        IS                          &$0.0033(\pm{0.014})$ &$0(\pm{0})$ \\ \hline
        IS(MDE)                     &$0.6797(\pm{0.3248})$    &$0.6470(\pm{0.3775})$ \\ \hline
        DRFS                        &$0(\pm{0})$ &$0(\pm{0})$ \\ \hline
        DRFS(MDE)                   &$0.0131(\pm{0.0252})$ &$0.0065(\pm{0.0190})$ \\ \hline
    \end{tabular}
    \caption{\label{tab:ijcai-2017-today-feats-distr}IJCAI-2017-Today-Users features distribution (* indicates significantly higher with p-value $< 0.05$ in Welch's two-tailed unpaired t-test).}
\end{table}

We find that, in CLPsych-2015-Users dataset, depression levels, such as ``Minimal'' and ``Mild,'' and  temporal patterns, such as  ``IS'' and ``DRFS'' are significantly higher in depressed population compared to the control population. Alternatively note that, instances labelled as ``None'' are significantly higher in the control population than in the depressed population. These distributions are expected, based on earlier research \cite{Choudhury2013Pred,kuppens2012emotional} and clinical criteria of depression \footnote{\url{https://www.psychiatry.org/patients-families/depression/what-is-depression}}.

In the IJCAI-2017-Ongoing-Users dataset, we note that the depressed population has a significantly higher Absence-Ratio 
than the control population. However, for all other features and in both IJCAI-2017-Ongoing-Users and IJCAI-2017-Today-Users datasets, we do not see any statistically significant difference. 





\section{Experimental Setup} \label{subsec:tud-exp-setup}

\begin{figure}
\centering
\includegraphics[width=1.01\textwidth]{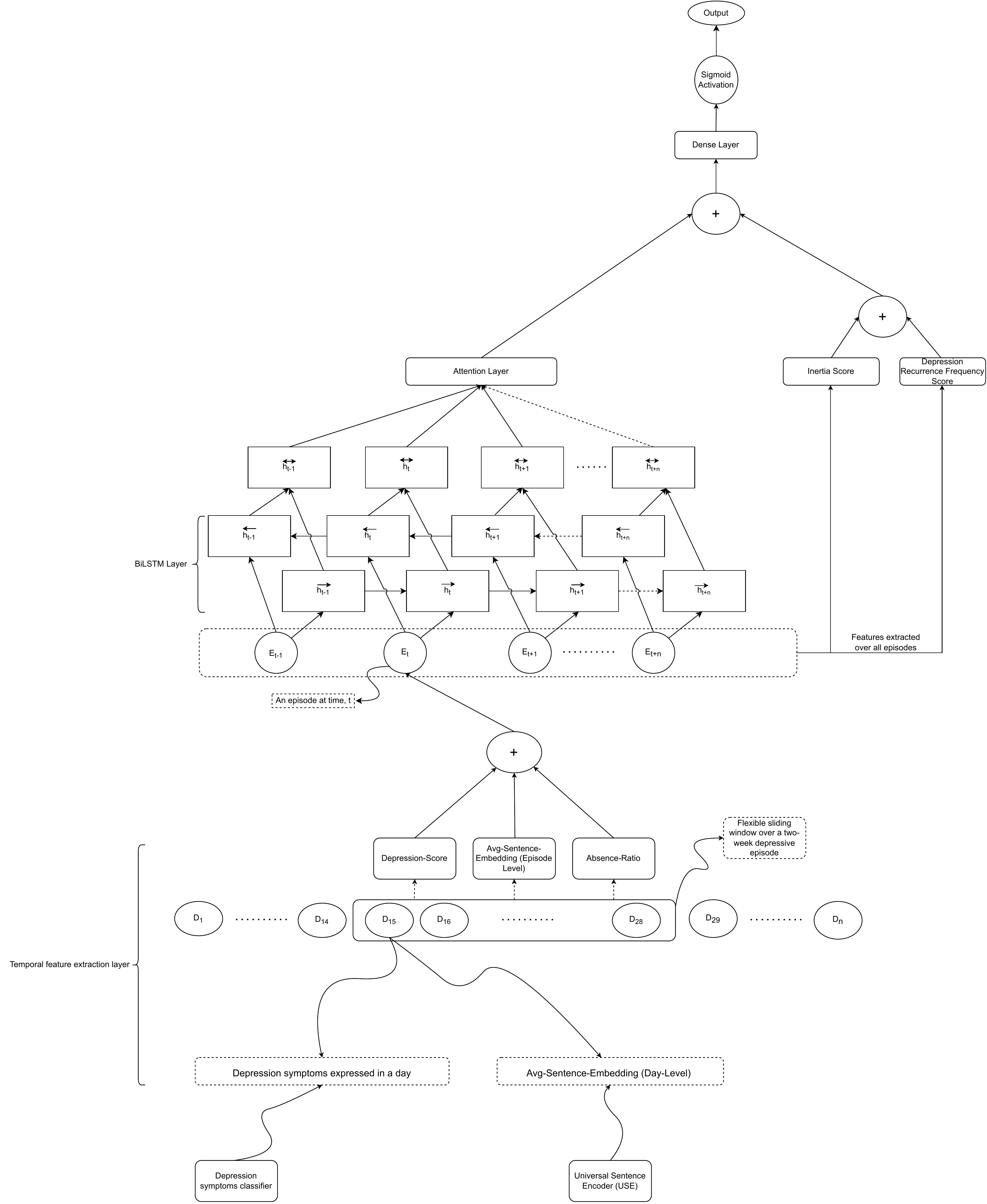}
\caption{\label{fig:detailed-tud}Detailed TUD model architecture (+ means concatenation, curved arrow followed by dashed box provides description of a component, solid arrow means data/process flow and dashed arrow means the same from ``n'' number of items).}
\end{figure}

Figure \ref{fig:detailed-tud} illustrates the over-all temporal deep-learning model. The model is provided with day-level aggregate depression scores and semantic representation of Tweets based on day-level average embedding. Subsequently, the model employs a flexible settings for different sliding day lengths and calculates episode level aggregates of depression score and semantic representations. This flexibility helps us doing three kinds of granular analysis over a user's depressive episodes based on slide length = 1, 7 and 14. These temporal episode level feature representations are later concatenated and further fed to a BiLSTM 
encoder to learn necessary temporal patterns of depression. This step produces an encoder output $h_{i}$ for each episode, and is further combined with final BiLSTM hidden representation, $h_{final}$. This is done for the entire temporal episode sequence to determine an attention weight, $w_{i}$ for each episode (Equations \ref{eqn:attention} and \ref{eqn:attn_weights}). Each $w_{i}$ is then normalized based on a softmax function which turns it to an attention score $\alpha_{i}$. This attention mechanism has been proposed by \cite{bahdanau2014neural} and is often called, ``Global Attention'' or ``Bahdanau Attention.'' Finally, we calculate a fixed length Attention score weighted sum of encoder outputs or episodes, $C$ (Equation \ref{eqn:weighted_sum}), which is further fed to a fully connected or dense layer followed by a sigmoid activation function which produces a binary value, ``1'' indicating presence or ''0'' absence of depression.  Hyperparameter settings for training TUD are provided in the Appendix \ref{appen:tud-train-params}. 

\begin{equation}
\label{eqn:attention}
    w_{i} = attention(h_{i}, h_{final})
\end{equation}

\begin{equation}
\label{eqn:attn_weights}
     \alpha_{i} = \frac{\exp(w_{i})}{\sum_{k=1}^{\abs{sequence}}\exp(w_{k})}
\end{equation}

\begin{equation}
\label{eqn:weighted_sum}
     C = \sum_{i=1}^{\abs{sequence}}\alpha_{i}h_{i}
\end{equation}

We report the accuracy scores (described in the next section) for user level depression detection task individually for each of our three datasets (i.e. CLPsych-2015-Users, IJCAI-2017-Ongoing-Users and Mixed-Users) for slide length=1 (because this provides us with the best results).


\begin{enumerate}
    \item \textbf{Ablation tests:}  We start with a model with all features (all-feats), then compare this model's accuracy scores with all the other feature ablated versions of it. 
    
    \item \textbf{Single feature tests:} We report the model's performance for each individual feature to asses that single feature's discriminatory power.
    
    \item \textbf{Best sliding length configurations:} We report the best model for other sliding configurations, i.e., for slide lengths = 7 and 14.
    
    \item \textbf{Baselines:} We create two baselines, such as: (1) \textbf{Episodic-Semantic-\\Representation based model (ES)}: this model uses ELSEA (Section \ref{subsec:tud-sem-info}) and a BiLSTM-Attention model 
    and (2)
    \textbf{All-Historic-Tweets-Semantic-Representation based model (HTS)}: this model uses ATEA representation (Section \ref{subsec:tud-sem-info}) followed by a fully connected layer for the binary depression detection task.
    
    \item \textbf{Non-Clinical vs Clinical Setting:} We also report whether following strict clinical criteria for depression detection, i.e., verifying the presence of either ``Anhedonia'' or ``Low Mood,'' makes any difference in user level depression detection compared to non-clinical settings (described in Section \ref{sec:tud-dep-score}).
    
    
    \item \textbf{Minimal Depression Expression (MDE) based temporal modelling:} Based on the depression level features distribution, we confirm that the ``None'' level is higher in control than depression (Section: \ref{sec:tud-clinical-feats-extract}), which indicates that we may try MDE to observe any increase the accuracy for DS. 
    
\end{enumerate}

\section{Evaluation}
\label{sec:tud-eval}
Since our task is a binary classification task, for accuracy analysis, we report Precision, Recall and F1 scores for each of our three datasets across the corresponding held-out sets and a test set. To enable 10 fold cross validation (CV), we create 10 (train set, held-out set) pairs. We then report average Precision, Recall and F1 scores and their standard deviations across this 10 folds. We also report, how our models trained on each folds do on a separate test set i.e. in IJCAI-2017-Today-Users dataset (Section: \ref{subsubsec:ijcai-2017-users}). 

This provides us with the information on how general our model is in a dataset with a totally different data distribution. We use two-tailed paired t-test and consider the difference between two accuracy scores as significant if p-value is $<0.05$.

\section{Results Analysis}
\label{sec:tud-results}

In this section we provide results analysis in the following dimensions (corresponding experiments are reported in Tables \ref{tab:acc-clpsych-2015-heldout}, \ref{tab:acc-clpsych-2015-test}, \ref{tab:acc-ijcai-2017-heldout}, \ref{tab:acc-ijcai-2017-test}, \ref{tab:acc-mixed-heldout} and \ref{tab:acc-mixed-test}). Underline indicates the score is significantly worse than that of the model which uses all the features (all-feats model). We perform pair-wise comparisons among our models and provide an analysis on the following dimensions:

\begin{table}
    \footnotesize
    \centering
    \begin{tabular}{|p{1.5cm}|p{1.5cm}|p{1.5cm}|p{1.5cm}|p{1.5cm}|p{1.5cm}|}
    \hline
        \textbf{Datasets} &\textbf{Category} &\textbf{Experiment-Name} &\textbf{Precision (Mean)} &\textbf{Recall (Mean)}  &\textbf{F1 (Mean)} \\ \hline
        CLPsych-2015-Users & feature ablation tests & all-feats & 0.7121 &0.7114 &0.7021 \\ \hline
        ~ & ~ & all-feats(MDE)  &0.7636	&0.6659	&0.7055	 \\ \hline
        ~ & ~ & - DS  &0.6749  &0.7317 &0.6815 \\ \hline
        ~ & ~ & - IS  &0.7253  &0.7615  &0.736  \\ \hline
        ~ & ~ & - DRFS        &0.7199 &0.7137 &0.7005 \\ \hline
        ~ & ~ & - TP &0.7405 &0.7289 &0.7230 \\ \hline
        ~ & ~ & - AR     &0.7176 &0.7718 &0.7371 \\ \hline
        ~ & ~ & - ES     &0.7649 &\underline{0.3521} &0.4671 \\ \hline
        ~ & ~ & ~ & ~ & ~ & ~ \\ \hline
        ~ & single features & DS  &0.6834 &0.3915 &0.467 \\ \hline
        ~ & ~ & DS(MDE) &0.623	&0.7142	&0.659 \\ \hline
        ~ & ~ & TP &0.6826 &0.4603 &0.4125 \\ \hline
        ~ & ~ & TP(MDE) &0.6248	&0.7214	&0.6592 \\ \hline
        ~ & ~ & AR &0.5327  &0.4520 &\underline{0.4593} \\ \hline
        ~ & ~ & ~ & ~ & ~ & ~ \\ \hline
        ~ & best slides & - DRFS (Slide-1) & See above & See above & See above \\ \hline
        ~ & best clinical settings & - AR(Slide-1) & See above  & See above & See above \\ \hline
        ~ & ~ & ~ & ~ & ~ & ~ \\ \hline
        ~ & baselines & (*) ES &0.6839  &0.7256 &0.6836 \\ \hline
        ~ & ~ & (+) HTS  &0.7057 &0.7485 &0.7091\\ \hline
        ~ & ~ & \cite{yadav2020identifying}  &- &- &0.7079\\ 
        \hline
    \end{tabular}
    \caption{CLPsysch-2015 dataset accuracy scores in held-out dataset\label{tab:acc-clpsych-2015-heldout}.}
\end{table}

\begin{table}
    \footnotesize
    \centering
    \begin{tabular}{|p{1.5cm}|p{1.5cm}|p{1.5cm}|p{1.5cm}|p{1.5cm}|p{1.5cm}|}
    \hline
        \textbf{Datasets} & \textbf{Category} & \textbf{Experiment-Name} & \textbf{Precision (Mean)} & \textbf{Recall (Mean)} & \textbf{F1 (Mean)} \\ \hline
         CLPsych-2015-Users & feature ablation tests & all-feats & 0.5358  &0.9611 &0.6872 \\ \hline
        ~ & ~ & all-feats(MDE)   &0.5218	&0.8166	&0.6363	\\ \hline
        ~ & ~ & - DS      &0.5093  &0.9722 &0.6659 \\ \hline
        ~ & ~ & - IS & 0.5302   &0.9944 &0.6913  \\ \hline
        ~ & ~ & - DRFS & 0.5509 &0.95 &0.6877 \\ \hline
        ~ & ~ & - TP     &0.5377 &0.9833 &0.6947 \\ \hline
        ~ & ~ & - AR         &0.5341 &0.9944 &0.6944 \\ \hline
        ~ & ~ & - ES & \underline{0} & \underline{0} & \underline{0} \\ \hline
        ~ & ~ & ~ & ~ & ~ & ~ \\ \hline
        ~ & single features & DS & \underline{0.0514} & \underline{0.1000}  & \underline{0.0679} \\ \hline
        ~ & ~ & DS(MDE) &\underline{0.4913}	&\underline{0.8000}	&\underline{0.6087} \\ \hline
        ~ & ~ & TP & \underline{0.1500}  & 0.3000  & \underline{0.2000} \\ \hline
        ~ & ~ & TP (MDE) &\underline{0.4764} &\underline{0.7833} &\underline{0.5924} \\ \hline
        ~ & ~ & AR & \underline{0} & \underline{0} & \underline{0} \\ \hline
        ~ & ~ & ~ & ~ & ~ &  ~ \\ \hline
        ~ & best slides & - AR (Slide-1) &See above &See above   &See above \\ \hline
        ~ & best clinical settings & - AR (Slide-1)(NCS) & 0.5368 & 0.9889 & 0.6953 \\ \hline
        ~ & ~ & ~ & ~ & ~ &  ~ \\ \hline
        ~ & baselines & (*) ES & 0.5101  & 0.9944  & 0.6742 \\ \hline
        ~ & ~ & (+) HTS & 0.5037 & 0.9944 & 0.6686 \\ \hline
       
    \end{tabular}
     \caption{CLPsych-2015-Users dataset accuracy scores in test dataset\label{tab:acc-clpsych-2015-test}.}   
\end{table}

\begin{table}
    \footnotesize
    \centering
    \begin{tabular}{|p{1.5cm}|p{1.5cm}|p{1.5cm}|p{1.5cm}|p{1.5cm}|p{1.5cm}|}
    \hline
        \textbf{Datasets} & \textbf{Category} & \textbf{Experiment-Name} & \textbf{Precision (Mean)} & \textbf{Recall (Mean)} & \textbf{F1 (Mean)}\\ \hline
       IJCAI-2017-Ongoing-Users & feature ablation tests & all-feats & 0.7669 & 0.7943 & 0.7770 \\ \hline
        ~ & ~ & all-feats(MDE) &0.7757	&0.7450	&0.7529	\\ \hline
        ~ & ~ & - DS & 0.7903 & 0.7766 & 0.7779\\ \hline
        ~ & ~ & - IS & 0.7722 & 0.7713 & 0.7703 \\ \hline
        ~ & ~ & - DRFS & 0.7594 & 0.7480 & 0.7502 \\ \hline
        ~ & ~ & - TP & 0.7513 & 0.7055 & 0.7200 \\ \hline
        ~ & ~ & - AR & 0.7662 & 0.6852 & 0.7174 \\ \hline
        ~ & ~ & - ES & 0.6263 & \underline{0.4487} & \underline{0.5031}\\ \hline
        ~ & ~ & ~ & ~ & ~ & ~ \\ \hline
        ~ & single features & DS & 0.5481 & 0.3002 & \underline{0.2854}\\ \hline
        ~ & ~ & DS(MDE) &\underline{0.5261}	&0.7893	&0.6259	\\ \hline
        ~ & ~ & TP & \underline{0.3647} & 0.6579 & 0.4317 \\ \hline
        ~ & ~ & TP(MDE) &\underline{0.5378}	&0.7890	&0.6297	\\ \hline
        ~ & ~ & AR & 0.6328 & 0.5690 & \underline{0.5863}\\ \hline
        ~ & ~ & ~ & ~ & ~ & ~ \\ \hline
        ~ & best slides & - DS(Slide-1) & See above &See above &See above \\ \hline
        ~ & best clinical settings & - DS(Slide-1) &See above &See above &See above \\ \hline
        ~ & ~ & ~ & ~ & ~ & ~ \\ \hline
        ~ & baselines & (*) ES & 0.7830 & 0.7746 & 0.7763 \\ \hline
        ~ & ~ & (+) HTS & 0.7447 & 0.7001 & 0.7138  \\ \hline
    \end{tabular}
    \caption{IJCAI-2017-Ongoing-Users dataset accuracy scores in held-out dataset\label{tab:acc-ijcai-2017-heldout}.}
\end{table}

\begin{table}
    \footnotesize
    \centering
    \begin{tabular}{|p{1.5cm}|p{1.5cm}|p{1.5cm}|p{1.5cm}|p{1.5cm}|p{1.5cm}|}
    \hline
        \textbf{Datasets} & \textbf{Category} & \textbf{Experiment-Name} & \textbf{Precision (Mean)} & \textbf{Recall (Mean)} & \textbf{F1 (Mean)} \\ \hline
       IJCAI-2017-Ongoing-Users & feature ablation tests & all-feats & 0.7632 & 0.8389 & 0.7975 \\ \hline
        ~ & ~ & all-feat(MDE) &0.7498 &0.7444 &0.7429 \\ \hline
        ~ & ~ & - DS & 0.7762 & 0.8444 & 0.8067 \\ \hline
        ~ & ~ & - IS & 0.7661  & 0.8278  & 0.7937 \\ \hline
        ~ & ~ & - DRFS & 0.7639 & 0.8222 & 0.7905  \\ \hline
        ~ & ~ & - TP & 0.7645 & 0.8056 & 0.7828\\ \hline
        ~ & ~ & - AR & 0.7543 & 0.6889 & 0.7129 \\ \hline
        ~ & ~ & - ES & 0.6535	& 0.5056	& 0.5613\\ \hline
        ~ & ~ & ~ & ~ & ~ & ~ \\ \hline
        ~ & single features & DS & \underline{0.1000} & \underline{0.2000} & \underline{0.1333}\\ \hline
        ~ & ~ & DS(MDE) &\underline{0.4976}	&\underline{0.7500}	&\underline{0.5964} \\ \hline
        ~ & ~ & TP & \underline{0.3529} & 0.7000 & 0.4692 \\ \hline
        ~ & ~ & TP(MDE) &\underline{0.4899}	&\underline{0.7166}	&\underline{0.5771} \\ \hline
        ~ & ~ & AR & \underline{0.6792} & \underline{0.6444} & \underline{0.6599}\\ \hline
        ~ & ~ & ~ & ~ & ~ & ~ \\ \hline
        ~ & best slides & - DS(Slide-1) & See above & See above & See above \\ \hline
        ~ & best clinical settings & - DS(Slide-1) & See above & See above & See above  \\ \hline
        ~ & ~ & ~ & ~ & ~ & ~ \\ \hline
        ~ & baselines & (*) ES & 0.7721 & 0.7611 & 0.7632  \\ \hline
        ~ & ~ & (+) HTS & 0.7473 & 0.7278 & 0.7317 \\ \hline
    \end{tabular}
    \caption{IJCAI-2017-Ongoing-Users dataset accuracy scores in test dataset\label{tab:acc-ijcai-2017-test}.}
\end{table}

\begin{table}
    \footnotesize
    \centering
    \begin{tabular}{|p{1.5cm}|p{1.5cm}|p{1.5cm}|p{1.5cm}|p{1.5cm}|p{1.5cm}|}
    \hline
        \textbf{Datasets} &\textbf{Category} &\textbf{Experiment-Name} &\textbf{Precision (Mean)}  &\textbf{Recall (Mean)} &\textbf{F1 (Mean)} \\ \hline
      Mixed-Users & feature ablation tests &all-feats &0.6249  & 0.7938 & 0.6951 \\ \hline
        ~ & ~ & all-feats(MDE) &0.7783	&0.6592	&0.7072 \\ \hline
        ~ & ~ & - DS & 0.6098 & 0.8315  & 0.7003 \\ \hline
        ~ & ~ & - IS & 0.6260  & 0.7078 & 0.6612 \\ \hline
        ~ & ~ & - DRFS & 0.6267  & 0.7930  & 0.6950 \\ \hline
        ~ & ~ & - TP & 0.6279 & 0.7783 & 0.6923 \\ \hline
        ~ & ~ & - AR & 0.6353 & 0.7745  & 0.6951 \\ \hline
        ~ & ~ & - ES & 0.6787  & \underline{0.3402} & \underline{0.4386} \\ \hline
        ~ & ~ & ~ & ~ & ~ & ~ \\ \hline
        ~ & single features & DS & 0.6728 & \underline{0.2793}& \underline{0.3852}\\ \hline
        ~ & ~ & DS(MDE) &0.5864	&0.7518	&0.6565 \\ \hline
        ~ & ~ & TP & 0.5725 & 0.5697 & 0.4395 \\ \hline
        ~ & ~ & TP(MDE) &0.5762	&0.7234	&0.6376  \\ \hline
        ~ & ~ & AR & 0.6223 & 0.4550 & 0.4840 \\ \hline
        ~ & ~ & ~ & ~ & ~ & ~ \\ \hline
        ~ & best slides & ES (Slide-7) & 0.6005 & 0.8579 & 0.7051 \\ \hline
        ~ & best clinical settings & HTS (Slide-1)(NCS) & 0.6957 & 0.7266 & 0.7071 \\ \hline
        ~ & ~ & ~ & ~ & ~ & ~ \\ \hline
        ~ & baselines & (*) ES & 0.6088  & 0.7984 & 0.6874 \\ \hline
        ~ & ~ & (+) HTS & 0.6935 & 0.7207 & 0.7037 \\ \hline
    \end{tabular}
    \caption{Mixed-Users dataset accuracy scores in held-out dataset\label{tab:acc-mixed-heldout}.}
\end{table}

\begin{table}
    \footnotesize
    \centering
    \begin{tabular}{|p{1.5cm}|p{1.5cm}|p{1.5cm}|p{1.5cm}|p{1.5cm}|p{1.5cm}|}
    \hline
        \textbf{Datasets} & \textbf{Category} & \textbf{Experiment-Name} & \textbf{Precision (Mean)} & \textbf{Recall (Mean)} & \textbf{F1 (Mean)}\\ \hline
      Mixed-Users & feature ablation tests & all-feats & 0.5906 & 0.9611 & 0.7305 \\ \hline
        ~ & ~ & all-feats(MDE) &0.7677	&0.6278	&0.6855 \\ \hline
        ~ & ~ & - DS & 0.5648 & 0.9556 & 0.7066 \\ \hline
        ~ & ~ & - IS & 0.6251 & 0.9167 & 0.7343 \\ \hline
        ~ & ~ & - DRFS & 0.5797 & 0.9556 & 0.7190 \\ \hline
        ~ & ~ & - TP & 0.5808 & 0.9556 & 0.7201\\ \hline
        ~ & ~ & - AR & 0.5992 & 0.9500 & 0.7317 \\ \hline
        ~ & ~ & - ES & \underline{0.0000}& \underline{0.0000}& \underline{0.0000}\\ \hline
        ~ & ~ & ~ & ~ & ~ & ~ \\ \hline
        ~ & single features & DS & \underline{0.0000} & \underline{0.0000} & \underline{0.0000} \\ \hline
        ~ & ~ & DS(MDE) &\underline{0.4862}	&\underline{0.7889}	&\underline{0.6016}\\ \hline
        ~ & ~ & TP & \underline{0.2500} & \underline{0.5000} & \underline{0.3333}\\ \hline
        ~ & ~ & TP(MDE) &\underline{0.4811}	&\underline{0.7667}	&\underline{0.5910} \\ \hline
        ~ & ~ & AR & \underline{0.0000} & \underline{0.0000} & \underline{0.0000} \\ \hline
        ~ & ~ & ~ & ~ & ~ & ~ \\ \hline
        ~ & best slides & HTS (Slide-1) & See below & See below & See below \\ \hline
        ~ & best clinical settings & HTS (Slide-1) & See below & See below & See below  \\ \hline
        ~ & ~ & ~ & ~ & ~ & ~ \\ \hline
        ~ & baselines & (*) ES & 0.5555 & 0.9889 & 0.7110 \\ \hline
        ~ & ~ & (+) HTS & 0.7583 & 0.8278 & 0.7904 \\ \hline
    \end{tabular}
    \caption{Mixed-Users dataset accuracy scores in test dataset\label{tab:acc-mixed-test}.}
\end{table}

\begin{enumerate}
    \item \textbf{Feature ablation study:} For all three dataset experiments, we do not see any significant accuracy difference among the ablated models and all-feats in both the held-out and test sets, except the avg-embedding ablated model, which performs significantly worse in majority of the cases.
    
    \item \textbf{Single feature study:} We report single features, i.e., depression-score (DS), absence-ratio (AR) and temporal patterns (TP) for all the experiment datasets. We use TP, which is a vector of two scores, i.e., IS and DRFS. TP (and so do IS and DRFS) is calculated over a user's timeline unlike a series of scores like DS and AR. As these scores are just single values, we do not believe IS and DRFS will produce any better predictive value than TP. We do not report their performance individually.  We see these models are highly unstable, i.e., they have high variability in accuracy scores across different folds in held-out and test sets. Performance becomes expectedly worse when the training and test sets are from different distributions, i.e., where number of episodes vary by a large margin (Tables \ref{tab:acc-clpsych-2015-heldout}, \ref{tab:acc-clpsych-2015-test}, \ref{tab:acc-ijcai-2017-heldout}, \ref{tab:acc-ijcai-2017-test}, \ref{tab:acc-mixed-heldout} and \ref{tab:acc-mixed-test}). 
    
    For datasets with the same data distribution (i.e., in held-out set of CLPsych-2015-Users and both held-out and test set for IJCAI-2017-Ongoing-Users), AR has the most predictive value.

    In general, it is clear that single features are data distribution sensitive, however, TP is a bit less sensitive compared to others. Moreover, DS has better performance in a dataset which has more depressive episodes compared to other one with fewer depressive episodes (Table \ref{tab:user-posting-stats}) and vice versa is true for TP. Except for AR in the IJCAI-2017-Ongoing-Users dataset, all the other single features perform poorly and under chance level. In addition, all the single features are significantly worse than the baseline and all ablated models,  except in some cases ES ablated model and all-feats models. We also note that DS performs worst for the model trained in CLPsych-2015-Users and tested in IJCAI-2017-Today-Users. Moreover, DS performance in the IJCAI-2017-Ongoing-Users is not as good as for the CLPsych-2015-Users. This could be because overall, the DS score is not a significantly important factor to discriminate between depressed and control populations in the IJCAI-2017-Ongoing/Today-Users dataset, as we have observed in Section \ref{sec:tud-clinical-feats-distr}.
    
    \item \textbf{Comparison with baseline models:} Both ES and HTS are significantly better than the avg-embedding ablated model and DS model across all datasets and in both held-out and test sets. In Mixed-Users test set, HTS is significantly better than all-feats and ablated models.
 
    Moreover, HTS is slightly (although not always significantly) better than ES in both held-out and test sets across all datasets except IJCAI-2017-Ongoing-Users dataset where the other way round is true. This signifies IJCAI-2017-Ongoing-Users dataset has more prominent temporal signals.  

     \item \textbf{Comparison with an early work baseline on same dataset} \cite{yadav2020identifying} used their fine-tuned DSD model to detect user-level depression in CLPsych-2015 dataset, which yielded $0.7079$ F1 accuracy (Table \ref{tab:acc-clpsych-2015-heldout}). Although, without more information it is hard to compare their accuracy with ours, we see that our least and best performing TUD models in all-feats and feature ablated models achieve $0.6815$ and $0.7371$ mean F1 accuracy which are only $\approx 2\%$ less and $\approx 3\%$ more than theirs, confirming the efficacy of our underlying DSD model for user-level clinical depression modelling (Table: \ref{tab:acc-clpsych-2015-heldout}).
    
    \item \textbf{Sliding lengths contribution:} Sliding length adjustments do not have statistically significant effect on accuracy gain for all our experiments, except in the Mixed-Users dataset where sliding length 14 has significantly worse performance.
    
    \item \textbf{Clinical vs non-clinical setting analysis:} Non-clinical setting does not have a statistically significant effect on accuracy gain compared to clinical setting, for all our experiments.

   \item \textbf{MDE based temporal modelling:} We observe statistically significant increase for DS and TP in this mode (i.e., DS(MDE) and TP(MDE)) for all the experiments in test sets (Tables \ref{tab:acc-clpsych-2015-test}, \ref{tab:acc-ijcai-2017-test} and \ref{tab:acc-mixed-test}). 
    We also find, instead of concatenating DS if we element wise multiply it with temporal embedding representation (ES) to create all-feats model (i.e., all-feats(MDE)), then there is some accuracy improvement over the original concatenation based all-feats model. This accuracy increase is also vetted by the statistically significant difference in ``None'', ``Minimal'' and ``IS'' level for this mode, where except IJCAI-2017-Today-Users dataset, we see ``None'' level is lower in depression population and higher in control population; ``Minimal'' and ``IS'' levels are higher in depression population and lower in control population in all other datasets (Tables \ref{tab:clpsych-2015-ongoing-feats-distr}, \ref{tab:ijcai-2017-ongoing-feats-distr} and \ref{tab:ijcai-2017-today-feats-distr}). In IJCAI-2017-Today-Users dataset, we see the statistically non-significant reverse distribution patterns for ``None'', ``Minimal'' and ``IS'' levels between these two populations. ``DRFS'' was found out to be majorly statistically non-significantly higher in depression population compared to control population for all datasets except IJCAI-2017-Today-Users dataset, where the other way round is true. It is interesting to see still in IJCAI-2017-Today-Users dataset the models perform well despite they were trained on a reverse distribution, this could be due to the fact that depression population have specific temporal pattern of depression scores which plays a discriminatory role here. We need to perform further investigation to confirm this in future.
    
    Compared to Non-MDE, single feature based models seem to be more stable in this mode across held-out and test sets for all the datasets (Tables \ref{tab:acc-clpsych-2015-heldout}, \ref{tab:acc-clpsych-2015-test}, \ref{tab:acc-ijcai-2017-heldout}, \ref{tab:acc-ijcai-2017-test}, \ref{tab:acc-mixed-heldout} and \ref{tab:acc-mixed-test}). However, through comparing the best models in each datasets in both held-out and test set for MDE and Non-MDE modes, we do not see their difference is statistically significantly different.  
    
    \item \textbf{Precision vs. recall:} We see that in held-out sets, precision and recall scores are close. However, in test sets, recall becomes higher and precision becomes lower, resulting in more sensitive classifiers. Change in training data distribution (i.e. trained in more temporal episodes) results in sensitive models (as evaluated in test data).
    
\end{enumerate}

All the above observations can be summarized into the following observations:

\begin{enumerate}
    \item Performance of DS depends on the dataset characteristic; if in a particular dataset, DS has significantly more discriminatory power then in that dataset DS might add more value.
    
    \item In general, single feature based models perform worse than all features and ablated features based models.
    
    \item Language only models (i.e., baseline models) are over-all pretty good in-terms of user-level depression detection, compared to posting behavior of the users, expressed depression in the posts through their depression scores and relevant temporal patterns. Note however, that those features can positively effect the model performance, provided that the data distribution is same in train and test sets. 

    \item Mixing two datasets with different distribution makes temporal modelling worse which is indicated by how HTS performs better than all other temporal models in the Mixed-Users dataset. Interestingly, making the depression score calculation more sensitive in MDE, we find DS becomes more effective as its accuracy increases significantly compared to the one that strictly follows the clinical thresholds. 
    
    \item Larger sliding length can result in similar model performance than more granular sliding lengths, indicating the promise for building more compact models in future.
\end{enumerate}

\section{Limitations} \label{sec:tud-limit}
Some limitations of our work are provided below:
\begin{enumerate}
    \item Our model uses depression score calculated based on the output of our Depression Symptoms Detection (DSD) classifier. This classifier is trained on a highly imbalanced dataset and is not robust to identify all the symptoms of depression from text. 

    \item We do not consider pure transformer models because earlier research do not indicate any extra benefit for this kind of temporal modelling. The amount of memory needed by the Self-attention \cite{vaswani2017attention} in a Transformer is quadratic on the length of the input, which means there is a significant limitation on the input size. Another shortcoming for using a transformer is that, to represent a sequence, an explicit mechanism to inform the classifier on the order of episodes is needed,  which is not necessary in our architecture. There is a state-of-art transformer model called Temporal Fusion Transformer (TFT) for temporal modelling \cite{lim2021temporal}, however, it is not yet established whether this TFT architecture is better. Interestingly, TFT has a close connection to our BiLSTM-Attention model in its model architecture. It is future work to consider other Attention mechanisms to see if there is any improvement.
    
    
    \item We strictly follow clinical criteria of depression detection, which limits us from experimenting with various lengths of depressive episodes, i.e., episodes larger than two weeks or less. Likewise, we emphasize on the expression of depression symptoms in a Tweet; if a candidate Tweet expresses depression but no particular symptom is detected (which is a rare possibility), that Tweet does not contribute to the depression scoring. 
    We do not explicitly account for other mental health conditions,  bereavement and other conditions that can resemble depressive symptoms. We also do not confirm whether any depressive symptom causes significant change in a user's daily life functioning. In future, we would like to investigate further to establish optimal thresholds and other depression criteria mentioned above in our clinical depression modelling.
    
    \item  Although LSTM might not perform good for longer sequence, BiLSTM followed by attention helps alleviate problems with longer sequence. 
    
    
    \item We largely follow the machine learning evaluation framework used in the seminal work of \cite{Choudhury2013Pred} for social media based depression detection; their sample size is also similar to us\footnote{However, they only reported average accuracy scores to compare different model-feature combinations. They did not report any standard deviation or t-test to compare their models. There is some chance of a Type-1 error in the paired t-test we used (i.e., it can reject the null hypothesis when it is actually true). One possible solution is to repeat cross-validation many times, which is extremely time and resource intensive for the deep learning models and therefore is not suggested \cite{dietterich1998approximate}. Another option is trying $5\times2$ fold CV; unfortunately, this has potential to largely decrease the training set size which may render deep learning models to be ineffective \cite{dietterich1998approximate}. Non-parametric significance test, such as, McNemer's test also has less statistical power in general, especially when used in small datasets like ours. We tried non-parametric pairwise Wilcoxon's signed ranked test. However, we did not see any difference of this test with our pair-wise student's t-test in terms of statistical significance finding.}. Moreover, most of the experiment results we report are not statistically significant, but paired t-test in 10 fold cross validation is robust against Type-2 error 
    \cite{dietterich1998approximate}, which means, when there is no significant difference between two model's accuracy, we can confidently assume that their accuracies are similar. We believe, our experiment results in an independent test set (i.e., IJCAI-2017-Today-Users) complement the analysis with the held-out set. Moreover, we find those clinical features have some discriminatory power which also have significant difference across depressed and control population (Section \ref{sec:tud-clinical-feats-distr}). This further corroborates the efficacy of our extracted clinical features. We have also focused on the nature of change in accuracy scores rather than comparing only their value, which also sheds light on the performance of our various model-feature combinations.

\end{enumerate}

\section{Conclusion} 
\label{sec:tud-conclusion}



We have described the construction of a deep temporal clinical depression modelling (TUD), using Twitter posts and all of their sub-components. These sub-components are created to help extract depression score (and few clinically relevant features based on it) from temporal social media posts. Later, we find their efficacy based on their accuracy for user-level depression detection when they are used with/out pure semantic embedding based temporal features in several modes of analysis. We observe that, clinical features are more useful in same data distribution and some of the features are dataset specific. Also, semantic embedding based representation is the most effective among all.

\begin{appendices}

\section{Depression Level Mapping}
\label{appen:depression-level-mapping}
In the following Table \ref{tab:depression-level-mapping}, we provide the mapping between the depression levels and corresponding range of depression scores. We use more stratification than the conventional PHQ-9 scale to get clearer idea about depression level distributions across our datasets.

\begin{table}[!h]
    \centering
    \begin{tabular}{|c|c|}
    \hline
            Depression Level &Depression Score Range\\
    \hline
            None                       & $0 - < 4$         \\
            Minimal                    & $4 - < 9$        \\
            Mild                       & $9 - < 14$        \\
            Moderate                   & $14 - < 19$        \\
            Moderately Severe          & $19 - < 27$       \\ 
            Severe                     & $ >= 27 $      \\
    \hline
    \end{tabular}
    \caption{\label{tab:depression-level-mapping} Depression level mapping reference.}
\end{table}

\section{Temporal User-level Clinical Depression Model (TUD) Training Configuration}
Here we report the training configuration for TUD (Table \ref{tab:tud-model-train-params})
\label{appen:tud-train-params}

\begin{table}[!htbp]
    \centering
    \begin{tabular}{|c|c|}
    \hline
            Hyperparameters &Values\\
    \hline
            \#Epochs        &10         \\
            \#Batch         &16         \\
             LSTM Hidden-Dimension &100 \\
            \#LSTM Hidden Layer &1 \\
            Drop-out &0.1 \\
            Learning Rate   &$1\times10^{-3}$          \\
            \#GPUs          &1          \\
            Loss function   &Binary Cross Entropy (BCE) Loss \\
            
    \hline
    \end{tabular}
    \caption{\label{tab:tud-model-train-params} TUD Model Hyperparameters.}
\end{table}

Since TUD is a binary classification task, we use the same settings for loss function as DPD described earlier.
\end{appendices}

\bibliography{refs} 

\end{document}